\documentclass{eceasst}

\usepackage{orcidlink}
\usepackage[english]{babel}
\usepackage{acronym}
\hypersetup{hidelinks}

\usepackage{placeins}
\usepackage{multirow}
\usepackage{booktabs}
\usepackage{siunitx}
\usepackage{comment}

\usepackage{subcaption}

\newif\ifforarxiv
\forarxivfalse 

\forarxivtrue 

\ifforarxiv
  \newcommand*\cold[1]{}
  
\else
\usepackage{lineno}
\linenumbers
\fi

\usepackage[accsupp]{axessibility}
\newcommand{\authorRef}[1]{\texorpdfstring{\autref{#1}}{}}
\newcommand{\authorOrcid}[1]{\texorpdfstring{\thinspace\orcidlink{#1}\thinspace}{}}

\usepackage{tikz}

\usetikzlibrary{arrows.meta,positioning,fit,backgrounds,calc}

\definecolor{slideblue}{HTML}{12385D}
\definecolor{accentred}{HTML}{EF3B24}
\definecolor{softred}{HTML}{FFF1EE}
\definecolor{teal}{HTML}{247C83}
\definecolor{softteal}{HTML}{EEF8F8}
\definecolor{steel}{HTML}{486A8F}
\definecolor{constraint}{HTML}{9B4639}
\definecolor{paper}{HTML}{F7F9FC}
\definecolor{linegray}{HTML}{D5DBE3}

\newlength{\redLaneY}
\newlength{\constraintRedGap}
\newlength{\topConstraintY}
\newenvironment{aiuse}{%
  \par\vspace{11pt}%
  \noindent\textbf{Declaration on the use of AI:}\enspace}{%
  \par\vspace{11pt}}

\newenvironment{data}{%
  \par\vspace{11pt}%
  \noindent\textbf{Data availability:}\enspace}{%
  \par\vspace{11pt}}

\title{Quantization Effects of Artificial Neural Networks for Embedded Edge-Computing Applications} 
\short{Quantization Effects of ANNs for Embedded Edge-Computing Applications} 
\author{
Alperen Aksoy\authorOrcid{0009-0001-3734-2157}\authorRef{1}, 
Ilja Bekman\authorOrcid{0000-0001-7562-3059}\authorRef{1}, 
Vesselin Dimitrov\authorOrcid{0009-0001-5158-3168}\authorRef{{1,2}}, 
Qader Dorosti\authorOrcid{0000-0001-9711-0609}\authorRef{2}, 
Chimezie Eguzo\authorOrcid{0000-0002-8286-101X}\authorRef{1}, 
Sarah Fleitmann\authorOrcid{0000-0002-7699-5708}\authorRef{1},  
Fabian Hader\authorOrcid{0000-0003-3490-3227}\authorRef{3}, 
André Zambanini\authorOrcid{0000-0002-1585-4397}\authorRef{1}, 
Stefan van Waasen\authorOrcid{0000-0003-0682-7941}\authorRef{{1, 4}}
} 
\institute{
Peter Grünberg Institute (PGI), Integrated Computing Architectures (ICA $|$ PGI-4), Forschungszentrum Jülich GmbH, Germany\autlabel{1}\\ 
Center for Particle Physics Siegen, Department für Physik, Universität Siegen, Germany\autlabel{2}\\ 
Faculty of Engineering, JARA-FIT Institute for Quantum Information, Forschungszentrum Jülich GmbH and RWTH Aachen University, Germany\autlabel{3}\\ 
University Duisburg-Essen, Germany\autlabel{4}
} 
\abstract{
This paper examines the use of \acp{QNN} for two resource-constrained scientific applications: automated calibration of semiconductor quantum bits (qubits) and scientific particle detectors. We evaluate the trade-offs between \ac{PTQ}, \ac{QAT}, and ultra-low-bit \acp{BNN} with respect to latency and resource usage. Our results demonstrate that \ac{PTQ} and \ac{QAT} are easily implementable solutions achieving a four-fold reduction in memory usage for \ac{U-Net} architectures, whereas \acp{BNN} are better suitable for use cases with challenging latency requirements.
For the training of non-differentiable custom \acp{BNN}, we propose a novel, hardware-constrained learning approach using \acp{GA}. We showcase a \ac{LUT}-based \ac{BNN} architecture suitable for direct conversion to \ac{VHDL} via the \acs{HCL4BNN} framework. This method achieves nanosecond-scale inference latencies at 10\,ns to 15\,ns without requiring specialized \ac{DSP} or \ac{BRAM} resources. 
} 


\begin{document}
\maketitle

\null
\newpage
\section*{Abbreviations}
\begin{acronym}[hcl4bnn]
    \acro{ADC}{Analog-to-Digital Converter}
    \acro{ANN}{Artificial Neural Network}
    \acro{BNN}{Binary Neural Network}
    \acro{BRAM}{Block RAM}
    \acro{CAM}{Content-Addressable Memory}
    \acro{CNN}{Convolutional Neural Network}
    \acro{DSP}{Digital Signal Processor}
    \acro{EAS}{Extensive Air Shower}
    \acro{FPGA}{Field Programmable Gate Array}
    \acro{eFPGA}{Embedded Field Programmable Gate Array}
    \acro{FF}{Flip-Flop}
    \acro{FP}{Floating Point}
    \acro{GA}{Genetic Algorithm}
    \acro{HDL}{Hardware Description Language}
    \acro{HLS}{High-Level Synthesis}
    \acro{HCL4BNN}{Hardware-Constrained Learning for Binary Neural Networks}
    \acro{LLM}{Large Language Model}
    \acro{LUT}{Look-Up Table}
    \acro{MAC}{Multiply-Accumulate}
    \acro{PTQ}{Post-Training Quantization}
    \acro{QAT}{Quantization-Aware Training}
    \acro{QNN}{Quantized Neural Network}
    \acro{ReLU}{Rectified Linear Unit}
    \acro{RFI}{Radio Frequency Interference}
    \acro{SiPM}{Silicon Photo-multiplier}
    \acro{U-Net}{U-shaped CNN} 
    \acro{VHDL}{Very High-Speed Integrated Circuit Hardware Description Language}
\end{acronym}
\null
\newpage


\section{Introduction}

Machine learning has proven to be an essential tool for the identification of patterns in complex datasets. A field, where this becomes increasingly important is experimental physics because it is dominated by measuring and processing data in large quantities. This leads to an increased use of machine learning algorithms for feature extraction.
As of now, it is mostly used in large data centers for offline data processing \cite{Kuznetsov2020}, but thanks to recent advancements of hardware and algorithms, it can be used more and more for edge processing applications \cite{Sengupta2025}. 
However, many deployment environments operate under severe energy, latency, and memory constraints \cite{Hader2025,Roessing2025,Dorosti2025,Augustin2025}, requiring dedicated efforts for processing hardware. 

\acfp{FPGA} (as well as embedded \acp{FPGA} or \acsp{eFPGA})\footnote{Some of the hardware-specific terms are explained in the appendix in \ref{sec:glossary}.} are frequently used for these edge computing tasks, since they offer low and deterministic latency and parallelism suitable to tackle the data rates of sensors in edge environments.
As processing units like CPUs and GPUs optimized for floating point operations differ fundamentally from \acp{FPGA}, full-precision neural networks cannot be executed verbatim on such resource-constrained platforms. 

To address this challenge, this work focuses on \acfp{QNN}, which reduce numerical precision to enable efficient edge computing while minimizing negative effects of the reduced dynamic range of values.

The choice of quantization strategy is heavily influenced by the specific constraints and processing demands of the application environment.
Therefore, we demonstrate the selected basic quantization methods for neural networks and their implementation for two distinct domains: 

\paragraph{Automated Qubit Tuning via \ac{PTQ} and \ac{QAT}} 
The precise calibration of semiconductor spin qubits requires detecting charge transitions in charge stability diagrams like in Figure \ref{fig:csd}. 
In order to automate this task, we utilize \ac{U-Net} architectures, a segmented \ac{CNN} consisting of a contracting and an expansive part \cite{Hader2025, Ronneberger2015}. 
The primary constraint in the physical setup is the limited cooling power within a cryostat, where heat dissipation from control electronics must be minimized. Ealier works show that \ac{U-Net}-like neural networks executed on specialized hardware accelerators would be compatible with the power budget available at the millikelvin to 4\,kelvin stages of cryostats used for qubit experiments \cite{hader_scalable_2025}.
We evaluate how \acf{PTQ} and \acf{QAT} affect the model performance in terms of detection quality of charge transitions in charge stability diagrams for \ac{U-Net} architectures with different numbers of parameters.

\paragraph{Real-Time Particle Detection via \acp{BNN}} 
In high-energy physics, detectors search for rare phenomena within a huge number of irrelevant events, hence generating massive data volumes that require immediate processing. Even though the experiments differ fundamentally in their research question, the detection technologies and thus data processing needs are often comparable.
In this work, we demonstrate data processing for the SHiP experiment at CERN \cite{Roessing2025}
. Real-time filtering there requires inference latencies in the micro- to nanosecond range, while a simple amplitude threshold cannot reliably separate relevant pulses from distorted or overlapping ones, so the trigger must make this decision itself.
Consequently, we focus on ultra-low-bit \acf{BNN}, trained via nature-inspired genetic optimization and translate the hardware constrained network directly to \ac{FPGA} fabric to achieve the necessary throughput and energy efficiency.

\paragraph{} To further investigate performance in terms of complexity trade-offs at the limit of numerical precision, we systematically evaluate the impact of extreme 1-bit quantization on the classification accuracy of highly compressed \acp{CNN}. 
Utilizing the MNIST dataset \cite{Lecun1998} as a benchmark, we analyze models trained through conventional methods --- distinct from the \acf{GA} approach used for hardware-native logic --- to demonstrate that carefully designed micro-architectures have comparable predictive performance.

\paragraph{Paper structure:} After an introduction to different paradigms of quantization of neural networks and a discussion of the used implementation methods, we present their application to mentioned examples.
This paper serves two purposes: first, it demonstrates the use of established \ac{PTQ}/\ac{QAT} strategies to a resource-constrained qubit-tuning task; second, it proposes a new method with a genetic-algorithm-trained, \acs{LUT}-native \ac{BNN} framework for a use case where the first methods are not sufficient.



\section{Quantization Strategies}
\label{sec:quantization}

This section summarizes the quantization strategies considered in this work and motivates their use in the two application domains mentioned above. Quantization maps high-precision floating-point values to lower-precision representations requiring fewer bits, ranging from 8-bit integers down to single-bit values. Depending on the bit width and model architecture, the resulting information loss may have little effect on task performance, or it may require adjusted training methods to recover accuracy.

\subsection{Post-Training Quantization} 
\ac{PTQ} serves as a strategy to convert a pre-trained floating-point model into a lower-precision representation without the requirement of retraining \cite{Jacob2018}. Within the field of experimental physics, particularly for the automated calibration of semiconductor spin qubits, \ac{PTQ} provides a rapid and practical pathway for model compression. This technique enables significant reductions in memory usage while maintaining a segmentation accuracy that is comparable to full-precision models. 
In favorable cases, \ac{PTQ} can preserve model performance making \ac{PTQ} a useful first step for deploying \acp{QNN} in energy-restricted edge environments. In this work, \ac{PTQ} is implemented directly in \texttt{PyTorch}  \cite{pytorch2019, pytorch2024}.

\subsection{Quantization-Aware Training} 
\ac{QAT} is a quantization strategy that simulates quantization effects during the training process itself by inserting fake quantization operators, like rounding, low resolution, or integer wrapping, into the model. For implementation of \ac{QAT} on top of \texttt{PyTorch} we use \texttt{QKeras} \cite{qkeras2021}, a framework providing quantized layer versions for deep neural network models.

This approach is designed to enhance the robustness of the \ac{QNN} against the precision loss typically associated with lower-bit representations \cite{Hubara2017}. 


\subsection{Extreme Quantization: Binary Neural Networks}
At extreme levels, \acp{BNN} constrain weights and activations to low-bit representations, thereby departing entirely from multi-level floating point training and inference. 
Although this generally requires a slightly larger network, it allows multiplications to be replaced by lightweight and fast logic operations and accumulations to become (unary) pop-counting procedures \cite{Rastegari2016}. This makes them an ideal candidate for \acf{LUT}-based \ac{FPGA} implementations.
Multiple approaches in recent years have addressed the challenge of training these structures \cite{Bacellar2025,Wang2019,Umuroglu2020,Weng2025}, since standard gradient-based methods are hindered by the non-differentiability of the binary design.

\paragraph{Training with Genetic Algorithms}

We are proposing an evolutionary optimization with \acp{GA} for the training of \ac{LUT}-based \acp{BNN}, bypassing the gradient requirement of standard backpropagation and making them directly applicable to binary-weight networks where gradients are undefined. \acp{GA} are nature-inspired optimization methods published in \cite{Holland1992} that mimic the processes behind the biological evolution \cite{Hornby2006, Meloni2016}.

Given a \texttt{population} of \texttt{individuals} (each representing a unique \ac{BNN}) with a set of \texttt{genes} (weights), the individuals' \texttt{fitness}-es (\ac{BNN} accuracy) are determined and the next \texttt{generation} (iteration) is constructed by \texttt{selecting} more successful individuals, \texttt{muta}\-\texttt{ting} (randomly changing few weights) and \texttt{crossing-over} (swapping sections of weights) their \texttt{genes}. 
Each new generation is evaluated again and consists by construction of more and more "fit" individuals. The condition for loop breakout may be that an individual achieves the target accuracy or that a certain number of iterations has been reached.

\paragraph{} With these three strategies established, the next section contrasts the first two and details how the third proposed \acp{BNN} architecture is mapped onto \ac{FPGA} hardware.

\section{Hardware Implementation and Deployment Flows}

This section summarizes the hardware deployment flows used in this work. We first describe conventional \ac{FPGA} workflows based on quantization and \ac{HLS}, and then introduce the proposed hardware-constrained approach for directly mapping \ac{LUT}-based \acp{BNN} to \acf{VHDL}.

\subsection{Conventional FPGA Workflows}

Conventional workflows for \ac{QNN} implementation in \acp{FPGA} typically develop from floating-point model design and training to quantization, optimization, and hardware synthesis as illustrated in Figure \ref{fig:QATvsHCL_A}. The steps in detail are: 

\textbf{(a)} Network definition in \ac{FP} representation, including input normalization;

\textbf{(b)} training via iterative back-propagation with weight adjustment using gradient-based optimization; 
quantization to map the \ac{FP} values to \ac{FPGA}-compatible fixed point or integer arithmetic; 
pruning and compression aiming to remove low impact connections and nodes; 

\textbf{(c)} expressing the necessary \ac{QNN} operations in synthesizable C++ and conversion to a hard\-ware-description language (\ac{HLS}), \textit{e.g.} via \texttt{Vitis\_HLS} \cite{UG1399_2025};

\textbf{(d)} integration into the target FPGA device firmware.

\vspace{-1em}
\begin{figure}[!ht]
    \centering
    \resizebox{\textwidth}{!}{%
\begin{tikzpicture}[ 
  font=\sffamily\small,
  >=Latex,
  block/.style={
    rounded corners=4pt,
    draw=slideblue!70,
    line width=0.75pt,
    fill=white,
    align=center,
    minimum height=0.92cm,
    inner xsep=5pt,
    inner ysep=4pt,
    text=slideblue
  },
  task/.style={
    block,
    fill=slideblue,
    draw=slideblue,
    text=white,
    text width=2.75cm,
    minimum height=1.45cm,
    font=\sffamily\bfseries
  },
  output/.style={
    task,
    text width=2.95cm,
    minimum height=1.45cm
  },
  redblock/.style={
    block,
    draw=accentred,
    fill=softred,
    text=accentred,
    font=\sffamily\bfseries\small
  },
  tealblock/.style={
    block,
    draw=teal,
    fill=softteal,
    text=teal,
    font=\sffamily\bfseries\small
  },
  chip/.style={
    rounded corners=8pt,
    draw=constraint!35,
    fill=white,
    line width=0.45pt,
    text=constraint,
    font=\sffamily\scriptsize\bfseries,
    inner xsep=6pt,
    inner ysep=3pt,
    align=center
  },
  frame/.style={
    rounded corners=6pt,
    draw=steel!45,
    line width=0.9pt,
    fill=paper,
    minimum width=12.65cm,
    minimum height=2.15cm
  },
  flow/.style={
    -{Latex[length=2.6mm,width=1.8mm]},
    line width=0.95pt,
    draw=slideblue
  },
  redflow/.style={flow, draw=accentred},
  tealflow/.style={flow, draw=teal}
]

\path[fill=white, draw=none] (-2.30,-3.80) rectangle (18.20,1.70);

\node[task] (detector) at (-0.40,0)
  {Detector task\\[-1pt]{\sffamily\scriptsize signals, backgrounds\\[-1pt]latency, rates}};

\node[frame] (codesign) at (8.20,0) {};


\node[redblock, text width=1.55cm] (py) at ([xshift=-4.75cm,yshift=\the\redLaneY]codesign.center)
  {Python\\model};
\node[redblock, text width=2.55cm] (qat) at ([xshift=-1.70cm,yshift=\the\redLaneY]codesign.center)
  {Hardware-aware\\training\\[-1pt]{\sffamily\scriptsize QAT, pruning\\[-1pt]reuse optimisation}};
\node[redblock, text width=2.00cm] (hls) at ([xshift=1.55cm,yshift=\the\redLaneY]codesign.center)
  {HLS\\mapping\\[-1pt]{\sffamily\scriptsize hls4ml / Vivado}};
\node[redblock, text width=1.65cm] (fwA) at ([xshift=4.75cm,yshift=\the\redLaneY]codesign.center)
  {Firmware\\entity};

\draw[redflow, shorten >=1.5pt, shorten <=1.5pt] (py) -- (qat);
\draw[redflow, shorten >=1.5pt, shorten <=1.5pt] (qat) -- (hls);
\draw[redflow, shorten >=1.5pt, shorten <=1.5pt] (hls) -- (fwA);





\node[output] (deploy) at (16.45,0)
  {Deployable\\inference system\\[-1pt]{\sffamily\scriptsize trigger / reduction\\[-1pt]firmware}};

\draw[flow, shorten >=1.0pt]
  (detector.east) -- ([xshift=-0.05cm]codesign.west);
\coordinate (redout) at ([xshift=-0.13cm,yshift=0.0cm]deploy.west);
\draw[redflow, shorten <=1.5pt, shorten >=1.5pt] (fwA.east) -- (redout);


\end{tikzpicture}
}%
\\[-5em]

    \caption{ Conventional workflow to deploy a \ac{QNN} on an \ac{FPGA} based on gradient training, quantization, and HLS conversion. }
    \label{fig:QATvsHCL_A}
\end{figure}

During inference, the floating point preprocessing and \ac{MAC}-heavy neuron evaluations can require significant latency and \acp{FPGA} resources, particularly \acfp{DSP} utilization.
Furthermore, step d) depends strongly on the optimization quality of the HLS conversion tool-chain with the relevant tools being introduced here.



\texttt{hls4ml} \cite{Fahim2021} is an open-source framework that converts trained machine learning models into \ac{FPGA} firmware using \ac{HLS}. It is designed to deploy neural networks with low latency and low power consumption, especially for real-time applications.

\texttt{FINN} \cite{Umuroglu2017} is an open-source experimental framework from AMD/Xilinx for accelerating the inference of \acp{QNN} on \acp{FPGA}. It focuses on generating highly optimized \ac{FPGA} implementations for low-precision networks such as binary and integer-quantized models.

\texttt{hls4ml} targets a broader set of machine-learning models and emphasizes simple, rapid conversion from frameworks such as \texttt{PyTorch} and \texttt{TensorFlow}~\cite{tensorflow2015} into \ac{FPGA} implementations. \texttt{FINN}, on the other hand, specializes in low-bit quantized neural networks and generates customizable \ac{FPGA} dataflow architectures.

In this work for \ac{BNN} baseline comparison both \acp{QAT} flows. 
First, we use the \texttt{Brevitas} library~\cite{Brevitas2024} with \texttt{PyTorch}, whose quantized models are exported to \texttt{FINN}. Second, we use \texttt{QKeras}~\cite{qkeras2021} for quantized layers in models converted by \texttt{hls4ml}.


These flows provide reference implementations against which the proposed \ac{LUT}-based \ac{BNN} approach is compared.

\subsection{Proposed Hardware-Constrained BNN Flow}
\subsubsection{Quantization Considerations}

To achieve the least latency for the signal processing at the inference time, we are realizing our network using \acp{LUT} on an \ac{FPGA} fabric, taking care to avoid \ac{DSP} or \acf{BRAM} clocked structures and seeking to use combinatorial logic, parts of which can be executed in sub-clock speed of the standard \ac{FPGA} fabric.

\begin{figure}[t]
    \centering\vspace{-5mm}
    \resizebox{\textwidth}{!}{%
\begin{tikzpicture}[ 
  font=\sffamily\small,
  >=Latex,
  block/.style={
    rounded corners=4pt,
    draw=slideblue!70,
    line width=0.75pt,
    fill=white,
    align=center,
    minimum height=0.92cm,
    inner xsep=5pt,
    inner ysep=4pt,
    text=slideblue
  },
  task/.style={
    block,
    fill=slideblue,
    draw=slideblue,
    text=white,
    text width=2.75cm,
    minimum height=1.45cm,
    font=\sffamily\bfseries
  },
  output/.style={
    task,
    text width=2.95cm,
    minimum height=1.45cm
  },
  redblock/.style={
    block,
    draw=accentred,
    fill=softred,
    text=accentred,
    font=\sffamily\bfseries\small
  },
  tealblock/.style={
    block,
    draw=teal,
    fill=softteal,
    text=teal,
    font=\sffamily\bfseries\small
  },
  chip/.style={
    rounded corners=8pt,
    draw=constraint!35,
    fill=white,
    line width=0.45pt,
    text=constraint,
    font=\sffamily\scriptsize\bfseries,
    inner xsep=6pt,
    inner ysep=3pt,
    align=center
  },
  frame/.style={
    rounded corners=6pt,
    draw=steel!45,
    line width=0.9pt,
    fill=paper,
    minimum width=12.65cm,
    minimum height=2.15cm
  },
  flow/.style={
    -{Latex[length=2.6mm,width=1.8mm]},
    line width=0.95pt,
    draw=slideblue
  },
  redflow/.style={flow, draw=accentred},
  tealflow/.style={flow, draw=teal}
]

\path[fill=white, draw=none] (-2.30,-3.80) rectangle (18.20,1.70);

\node[task] (detector) at (-0.40,0)
  {Detector task\\[-1pt]{\sffamily\scriptsize signals, backgrounds\\[-1pt]latency, rates}};

\node[frame] (codesign) at (8.20,0) {};





\node[tealblock, text width=1.55cm] (bin) at ([xshift=-4.75cm,yshift=0cm]codesign.center)
  {INT / BIN Python\\model};
\node[tealblock, text width=2.55cm] (ga) at ([xshift=-1.70cm,yshift=0cm]codesign.center)
  {GA /\\architecture search\\[-1pt]{\sffamily\scriptsize forward pass}};
\node[tealblock, text width=2.00cm] (hdl) at ([xshift=1.55cm,yshift=0cm]codesign.center)
  {FPGA-native\\HDL\\structure};
\node[tealblock, text width=1.65cm] (fwB) at ([xshift=4.75cm,yshift=0cm]codesign.center)
  {Firmware\\entity};

\draw[tealflow, shorten >=1.5pt, shorten <=1.5pt] (bin) -- (ga);
\draw[tealflow, shorten >=1.5pt, shorten <=1.5pt] (ga) -- (hdl);
\draw[tealflow, shorten >=1.5pt, shorten <=1.5pt] (hdl) -- (fwB);


\node[output] (deploy) at (16.45,0)
  {Deployable\\inference system\\[-1pt]{\sffamily\scriptsize trigger / reduction\\[-1pt]firmware}};

\draw[flow, shorten >=1.0pt]
  (detector.east) -- ([xshift=-0.05cm]codesign.west);
\coordinate (tealout) at ([xshift=-0.13cm,yshift=-0cm]deploy.west);
\draw[tealflow, shorten <=1.5pt, shorten >=1.5pt] (fwB.east) -- (tealout);


\end{tikzpicture}
}%
\\[-5em]

    \caption{ Proposed hardware-constrained learning approach with direct HDL generation.}
    \label{fig:QATvsHCL_B}\vspace{-3mm}
\end{figure}

We propose a hardware constrained approach: to design the binary network using only \ac{FPGA}-appropriate operations implementable with \ac{LUT}-building blocks (\textit{e.g.}\ \ac{CAM} or adders), through routing (\textit{e.g.}\ division by powers of 2, \textit{i.e.}\ bit shifting), or other combinatorial logic (\textit{e.g.}\ carry chains, multiplexers) that avoid clocked structures as in \cite{Bacellar2025}. 
This constraint eliminates the quantization and conversion steps but requires training suitable for non-differentiable operations.

Rather than approximating conventional floating-point multiplication, the proposed architecture replaces arithmetic neuron operations with a constrained set of \ac{FPGA}-native logical transformations that are optimized for low-resource inference.


Unlike conventional 1-bit binary weights, which can only encode a sign flip (\textit{e.g.}, \{+1,-1\}), our synapse model requires 2 bits per weight for four distinct operations: blocking, passing through unchanged, increasing, and negating the input. 
This choice preserves the coarse, non-arithmetic character of binary-style weights while adding the "Block" state needed for the sparsity objective of the genetic algorithm (Sec.~\ref{sec:ga_train}) and the "Pass" state that avoids unnecessary sign inversion. 
 
The same reasoning extends to the choice of 2-bit neuron values (rather than 1-bit): four activation levels allow a neuron to be selectively increased or decreased rather than only switched on/off, while remaining small enough for efficient \ac{LUT} mapping, although higher bit widths are usable as well and are used for multi-bit input for example (see below).

\begin{itemize}
\itemsep-0.3em 
    \item $w=0 \Rightarrow$ \texttt{Block}: Blocking operation, output is set to $0$ regardless of input, which auto-prunes this synapse (also see \ref{sec:ga_train}).
    \item $w=1 \Rightarrow$ \texttt{Pass}: Passing the input value through unchanged.
    \item $w=2 \Rightarrow$ \texttt{Incr.}: Increasing the input value, settling on binary shift left ($ < < 1$) with saturation safeguard (limiting to respective \texttt{MAX\_INT}). 
    \item $w=3 \Rightarrow$ \texttt{Neg.}: Negation of the value, which is represented by bit-wise inversion, avoiding classical $+1$ correction for the implementation efficiency.
\end{itemize}

This allows a multiplication operation to be replaced by a very hardware-efficient \ac{LUT}-operation in a $4 \times 4$ \ac{CAM}, illustrated in the following Eq. \ref{eq:CAM}, fitting efficiently into a single standard LUT4/LUT6~\cite{Milenkovic2024} primitive:

\vspace{-1em}
\begin{equation}
\begin{array}{llll|cl|llll} 
    \multicolumn{4}{c}{\text{2-bit  input}} & \multicolumn{2}{c}{\text{weight}} & \multicolumn{4}{c}{\text{2-bit output}} \\
    0 & 1 & 2 & 3 & w=0 & \text{Block} & 0 & 0 & 0 & 0 \\
    0 & 1 & 2 & 3 & w=1 & \text{Pass} & 0 & 1 & 2 & 3 \\
    0 & 1 & 2 & 3 & w=2 & \text{Incr.} & 1 & 2 & 3 & 3 \\
    0 & 1 & 2 & 3 & w=3 & \text{Neg.} & 3 & 2 & 1 & 0 
\end{array}
\label{eq:CAM}
\end{equation}

While this is viable for the hidden layers, the input layer in edge applications is often connected to multi-bit sensor outputs, \textit{e.g.}\ 12-bit \ac{ADC} samples, so it is most useful to use integer values.
The network input is reduced without normalization from 12\,bit to 7\,bit, which is a trivial operation in an \ac{FPGA}.

We adapt the activation function to represent one of the four operations at inference time:

\vspace{-1em}
\begin{equation}
\begin{array}{l|cl|l} 
    \multicolumn{1}{c}{\text{INT input}} & \multicolumn{2}{c}{\text{weight}} & \text{INT output} \\
    v & w=0 & \text{Block} & 0 \\
    v & w=1 & \text{Pass}  & v \\
    v & w=2 & \text{Incr.} & \min(v < < 1, \text{MAX\_INT}) \\
    v & w=3 & \text{Neg.} & \sim v 
\end{array}
\label{eq:CAM_int}
\end{equation}

An integer summation of $N$ weighted inputs per neuron is performed, allowing for the sum bit width of at least $\log_2(3N)$ to prevent an overflow. 
This step introduces a long carry chain, which is depending on the layout and is optimized in hardware by pairwise tree addition, \textit{e.g.} {\small $((1+2)+(3+4))$}. This reduces the carry chain length logarithmically while maintaining a fully combinatorial implementation without pipeline stages.
The network structure is constrained to $2^n$ neurons per layer for efficient summation.

We use an activation function inspired by a \ac{ReLU}~\cite{Agarap2019}, which maps the integer sum back to 2-bit neuron values using three thresholds to separate the sum into four bins.
As our network is not considering biasing, the thresholds are calculated during training based on the number of inputs to the neuron, as more inputs yield higher accumulation values. During training, some of the weights to these inputs might be set to zero (\texttt{Block}), effectively deactivating this input. Hence, this needs to be taken care for the threshold values as well. This way, the full range of the activation is available also to neurons, which inputs turn out to be heavily pruned.

The \ac{ReLU}-like quantized "staircase" activation function uses the discrimination with three fixed thresholds. 
Those are set at training time, depend on number of non-zero weights going into the neuron, and act directly on integer sums to avoid normalization step at runtime.

To give an example of the activation function for four non-zero neurons, we need to divide the maximum sum of $3\cdot 4$ into 4 output values $0$, $1$, $2$, $3$ via inclusive thresholds: 2, 6, 10.

The final network output is threshold encoded, with neuron values 0 and 1 as "off" and 2 and 3 as "on". 
The proposed approach assumes that the target classification tasks remain separable under ultra-low-bit representations and coarse logical activation transformations.

\subsubsection{Training with \aclp{GA} \label{sec:ga_train}}
To implement the \ac{GA}, we have used the \texttt{deap} Python package~\cite{DEAP2012} and its \texttt{eaSimple} procedure is derived to include \texttt{elitism}, which transfers the $n$ best individuals unchanged to the next generation. 
This accommodates for the stochastic dips of the noisy fitness function described in more detail in the following.

\begin{figure}[!ht]
    \centering
    \includegraphics[width=0.495\linewidth]{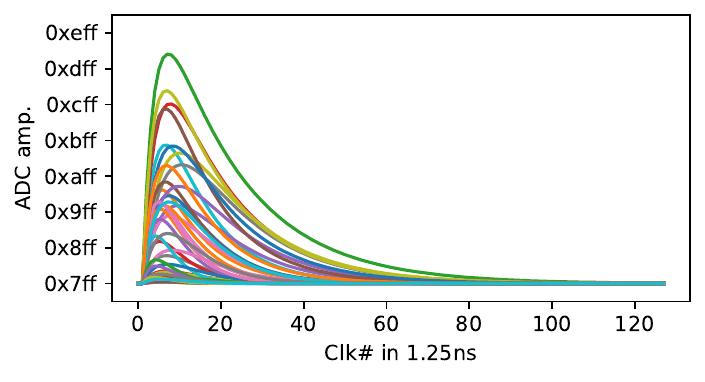}
    \includegraphics[width=0.495\linewidth]{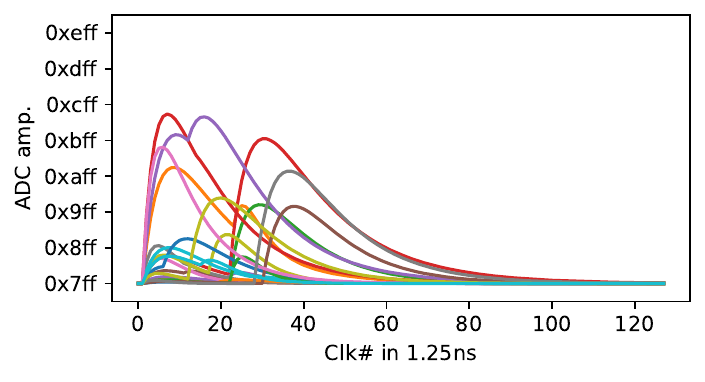}

    \caption{
    Left: isolated simulated SiPM pulses ("Good" class). Right: two overlapping or distorted simulated SiPM pulses ("Ugly" class). }
    \label{fig:SiPM_Sim}
\end{figure}

\paragraph{
Fitness Evaluation
}
For training, reference data is required. In the case of the \ac{SiPM} readout in the application described in section \ref{sec:res-sipm}, we use empirical double exponential functions. All SiPM training and evaluation data was generated by \texttt{HCL4BNN}~\cite{HCL4BNN} using the \texttt{SiPMDataset} Python class. The relevant distinction criterion is clean waveform based on a single input pulse ("good") versus distorted waveforms with multiple inputs ("ugly"), as depicted in Figure \ref{fig:SiPM_Sim}.
Alternatively, dedicated simulation frameworks may also be used~\cite{Rodriguez2025}.

For each individual a set of typically 300 "good" and 300 "ugly" waveforms are generated. The \ac{BNN} predicts tuples: a two-element binary class vector with $(1,0)$ for "good" and $(0,1)$ for "ugly". Due to both outputs being calculated for themselves, a consequential classification of neither "good" nor "ugly" would be encoded as $(1,1)$ or $(0,0)$, respectively. This can be used to indicate an abstention.
Since the training set is regenerated for each evaluation, the fitness exhibits erratic behavior. The impact is discussed further in \ref{sec:static_regen} together with the results.

The accuracy score measures how well the predicted tuples match the target tuples, assigning credits for partially correct predictions, as shown in Eq.~\ref{eq:fitness} for tuples (left) and MNIST targets (right). It is then normalized by the tuple length and size of the training set. To avoid trivial "broken clock" classifiers from achieving 50\,\% accuracy by always predicting the same class, goodness is set to 0\,\% whenever all predictions are identical. Otherwise, training stagnates, since complex models in early stages cannot out-compete the trivial variants in the same generation prior to further optimization and are selected for reproduction less frequently.

\vspace{-1em}
\begin{equation}
\arraycolsep=8pt
\begin{array}{ccc} 
    \multicolumn{3}{l}{\text{for tuple target = } (1,0)} \\
    (1,0) & \hat{=} & 1.0 \\
    (1,1) & \hat{=} & 0.5 \\
    (0,0) & \hat{=} & 0.5 \\
    (0,1) & \hat{=} & 0.0 \\
\end{array}
\quad
\quad
\quad
\begin{array}{ccc} 
    \multicolumn{3}{l}{\text{with digits 0, 1, 2, 3, 4, 5, 6, 7, 8, 9}} \\    
    \multicolumn{3}{l}{\text{for MNIST target = } (0,0,0,0,0,0,0,0,0,1)} \\
    (0,0,0,0,0,0,0,0,0,1) & \hat{=} & 10/10 \\
    (0,0,0,0,0,0,1,0,0,1) & \hat{=} & 9/10 \\
    \\
\end{array}
\label{eq:fitness}
\end{equation}

When calculating the fitness of an individual $f$ we see an opportunity to combine two objectives of higher accuracy $a$ and more sparse i.e. smaller network, measured by the fraction of zero-weights $w_\text{zero}$.
Rather than performing a Pareto-based multi-objective optimization, we use a weighted sum of these two criteria with scales of 10 and 1 respectively. 

The resulting composite fitness can be maximized within \texttt{deap}~\cite{Konak2006} as:

\vspace{-1em}
\begin{equation}
f = a\cdot10 + \frac{w_\text{z} }{w_\text{tot}} \cdot 1
\label{eq:fitness_weighted}
\end{equation}

\begin{figure}[t]
    \centering
    \includegraphics[width=0.49\linewidth]{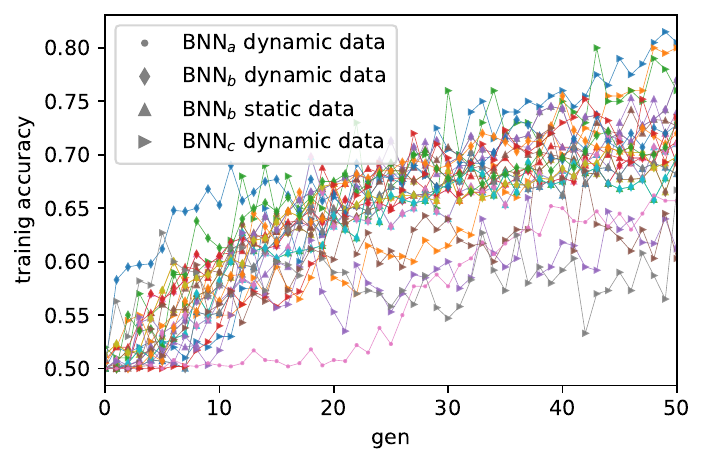} 
    \includegraphics[width=0.49\linewidth]{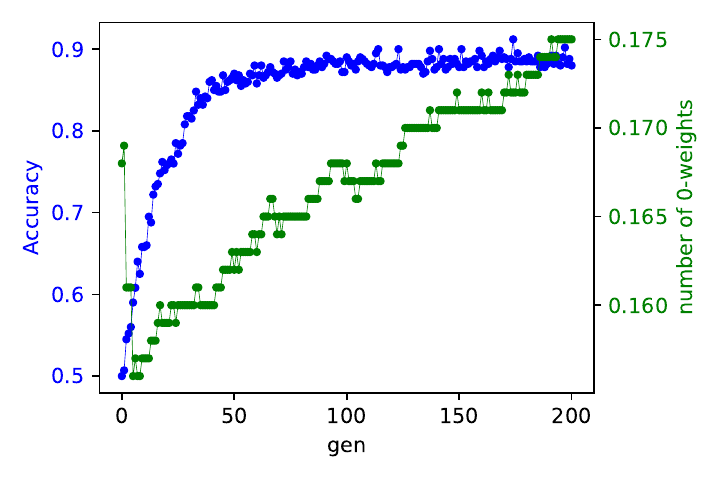}

    \caption{
    Left: Best training accuracy progression 
    across generations for several \ac{GA} runs, see Table \ref{tab:results} for validation results.
    Right: Progression of best accuracy and network sparsity (measured by the number of zero-weights)  across generations for a longer \ac{GA} run.}
    \label{fig:ga:training}
\end{figure}

The relative scales of 10 and 1 were chosen to reflect the typically logarithmic shape of the accuracy improvement over generations (see Figure \ref{fig:ga:training} on the right): accuracy tends to increase rapidly in early generations before entering a slow, near-stagnant phase. 
Weighting accuracy an order of magnitude above the sparsity term ensures that the composite fitness remains dominated by $a$ and the search is not diverted toward smaller but less accurate individuals. 
In the later slow-growth phase, individuals of comparable accuracy are then further differentiated by the sparsity term, favoring smaller networks among near-equal performers. 
The specific values were not individually tuned; a decade of separation was chosen to establish this priority ordering robustly rather than to optimize the trade-off precisely. The training is typically performed using populations above 300 individuals or even 1000 individuals if computation allows.

\FloatBarrier

\subsubsection{\acl{HDL} Conversion and Inference\label{sec:hw-hdl}}


Our Python code generates directly usable \ac{VHDL} \texttt{entity} code from the weights (\ac{LUT}-con\-stants) and sum threshold constants, using the results of the training.
A custom \ac{VHDL} \texttt{package} supports this step by defining functions with operator overloading, making the \ac{HDL} from Python output easier to read. 


The \ac{FPGA} implementation software can optimize across layers and neurons so that further compression of up to 20\,\% of the network footprint can be achieved without functional changes, \textit{e.g.}\ via \ac{LUT} packing \cite{UG904_2025}. 
Optionally, the \ac{VHDL} \texttt{attribute} \texttt{KEEP}~\cite{UG901_2025} can be used to keep the layers separated for ease of debugging.

\paragraph{}
In contrast to the related approaches \texttt{LUTNet}~\cite{Wang2019} and \texttt{LogicNets}~\cite{Umuroglu2020}, which also map trained network components directly onto \acp{FPGA}, our approach does not rely on gradient-based training and differentiable synapse behavior. This makes our \ac{GA}-based approach applicable to explicitly hardware-mapped, non-differentiable structures without needing a differentiable surrogate for them. The resulting training pipeline becomes simpler and does not depend on backpropagation infrastructure like \texttt{PyTorch} or \texttt{TensorFlow}.

\paragraph{}
Having established the quantization paths, the next section applies them to two use cases from experimental physics with varying constraints.

\clearpage
\section{Domain-Specific Evaluation and Results}


This section evaluates the quantization strategies introduced above in two application domains. 

\subsection{Motivation for Quantization}
The first case study on segmentation for automated qubit tuning focuses on an energy-efficient implementation of a neural network. When porting to \ac{FPGA} hardware, this directly leads to memory-efficiency. As the latency requirements are secondary, only conventional \ac{PTQ} and \ac{QAT} are investigated.

The second case study on signal classification for particle detectors requires ultra-low-latency data processing as new data is continuously taken with high rates and proceeding processing steps need to be activated. Thus, the more agressive \ac{BNN} inference is pursued.

\subsection{Memory-Efficient Qubit Tuning using PTQ and QAT Analysis\label{sec:res-unets}}

Our research on the trade-offs between quantization strategies and detection quality evaluated different \ac{U-Net} architectures with varying parameter counts: the compact UNet-38k and the extremely lightweight UNet-447, both architectures described in \cite{hader_scalable_2025}.

\begin{figure}[ht]
    \centering
    \begin{subfigure}{0.4\linewidth}
        \includegraphics[width=\linewidth]{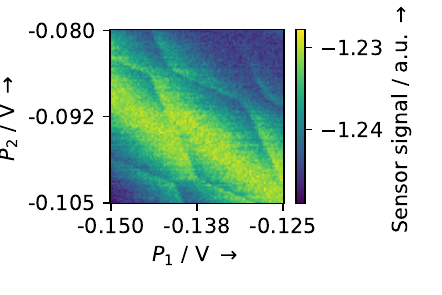}
        \subcaption{Without sensor compensation}
        \label{fig:uncompensated_csd}
    \end{subfigure}
    \begin{subfigure}{0.4\linewidth}
        \includegraphics[width=\linewidth]{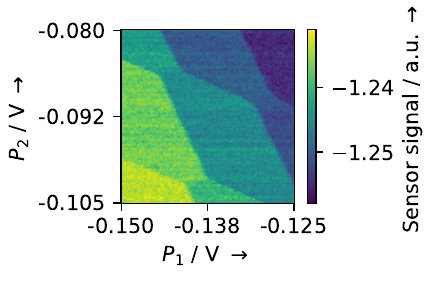}
        \subcaption{With sensor compensation}
        \label{fig:compensated_csd}
    \end{subfigure}
    
    \caption{Simulated charge stability diagram with and without compensation of the sensor response to the voltage changes for gates $P_1$ and $P_2$ \cite{hader_scalable_2025}. Further data examples can be found in \cite{5PB3GT_2024}.}
    \label{fig:csd}
\end{figure}

The inputs for the \ac{U-Net} are so called charge stability diagrams similar to Figure \ref{fig:csd}, where a charge sensor signal is shown in dependency on the gate voltages of the qubit structure. The processing task with these diagrams is to identify transitions between regions, \textit{i.e.}\ identifying the edges. To be able to use lightweight neural networks for the detection of the charge transition lines in these diagrams, the unwanted response of the sensor to the change of the applied voltages is first compensated like shown in Figure \ref{fig:compensated_csd}. In the following, compensated charge stability diagrams are used and simulated with the help of SimCATS~\cite{simcats}.

The results presented in Table~\ref{tab:ptq_vs_qat} are achieved with the help of the quantization module of \texttt{PyTorch} following the tutorial in \cite{ferianc_quantization}. A description of the training process and the used hyperparameters is available in appendix \ref{app:unet_hyperparameters}. The loss function used for training is the sum of the binary cross entropy~\cite{bce_pytorch} and the dice similarity coefficient~\cite{dice_score}, which balances overall overlap of ground truth and predicted mask with pixel-wise accuracy. During training a dataset with \num{10000} randomly sampled SimCATS parameter configurations and 100 charge stability diagrams per configuration is used \cite{simcats_datasets}. The performance evaluation is done with the help of a separate test set which was generated by using \num{1000} randomly sampled SimCATS parameter configurations and the dice similarity coefficient as accuracy metric.\\
The results demonstrate that \ac{PTQ} serves as a rapid and practical pathway for model compression, achieving significant memory reduction while maintaining segmentation accuracy comparable to full-precision models. The results achieved for \ac{QAT} are only marginally better when looking at the mean dice similarity coefficient. However, differences can be seen in the standard deviation, where \ac{QAT} achieves a similar standard deviation as the unquantized models and \ac{PTQ} has a slightly higher standard deviation. Performance differences for \ac{PTQ} and \ac{QAT} between the two \ac{U-Net} models could not be identified.

By reducing the numerical precision of weights and activations from 32-bit floating-point to 8-bit integers, we can reduce memory consumption $4 \times$ and computational cost $16 \times$ when the models are executed on appropriate hardware, for instance \acp{FPGA}.



\begin{table}[!ht]
    \centering
    \begin{tabular}{c|c|c|c}
         Model & Unquantized & \ac{PTQ} & \ac{QAT}  \\
               & in \% & in \% & in \%  \\
         UNet-447 & $85 \pm 28$ & $83 \pm 29$ & $85 \pm 28$ \\
         UNet-38k & $96  \pm 9$ & $96 \pm 9$ & $96 \pm 9$
    \end{tabular}
    
    \caption{Mean dice similarity on the test set with standard deviation on the test set.}
\label{tab:ptq_vs_qat}
\end{table}

\FloatBarrier

\subsection{Ultra-Low-Bit \ac{BNN} Evaluation for Particle Detector Signal Classification}



\subsubsection{Preliminary MNIST Benchmark}
We used MNIST as a compact benchmark to test the performance vs. complexity trade-off of highly compressed binary networks, before applying the \ac{BNN} to the detector signals. 

This benchmark is not intended to model particle-detector data, but to provide a simple reference for how much classification accuracy can be retained with very small binary architectures. 
To systematically evaluate performance-complexity trade-offs, we trained highly compressed \acp{CNN} with binary weights and activations on the MNIST dataset~\cite{Lecun1998} containing hand-written digits. While a better benchmark would be another time-series dataset, MNIST offers well-known accuracy target.

\begin{figure}[ht]
    \centering
    \includegraphics[width=1.0\linewidth]{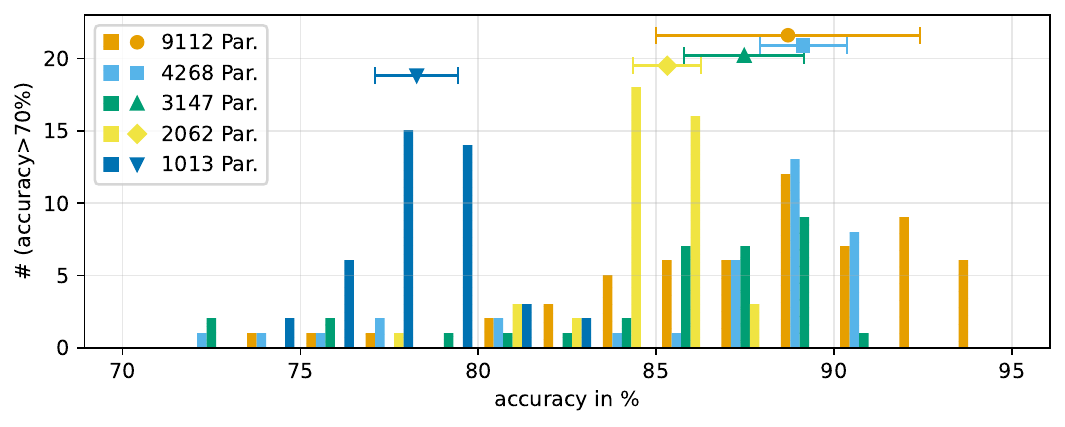}
    \caption{ Illustration of the accuracy distributions of 5 compressed \acp{CNN} (next to each other per accuracy bin) for training reruns on MNIST dataset after 10 epochs. Largest CNN (orange color, circle) distribution has the mean at $89\%$, 
    Smallest CNN (purple color, down-triangle) peaks at $78\%$. Means and standard deviation points are calculated over the normal part of the distributions.
    }
    \label{fig:mnist_cnns}
\end{figure}

Models with as few as 1013 binary parameters achieved an accuracy above 70\,\% on CPU for the classification of the MNIST-digits, illustrating that carefully designed micro-architectures can preserve performance even at extreme compression levels (see Table \ref{tab:mnist_results} and Figure \ref{fig:mnist_cnns}).
Because some training runs may get stuck in local minima, we use an accuracy threshold to identify runs that should be restarted. Specifically, a run is considered to have become stuck if its accuracy is below $70\,\%$ after 10 epochs.
Without the knowledge of absolute reachable accuracy for a given model a near-normal distribution of converged runs is considered as estimate and so not the highest accuracy but the mean over converged runs' accuracies is taken. 
The network with the higher number of parameters is able to achieve higher accuracy, although on average a larger spread indicates difficulty to converge to its global optimum.
Study of convergence behavior and especially comparison to the genetic algorithm has to be performed at a future point. 

\begin{table}[!h]
    \centering
    \begin{tabular}{r | ccccc}
        \# Parameters & 9,112 & 4,268 & 3,147 & 2,062 & 1,013 \\
        Accuracy in \%      & $88.7\pm 3.7$ & $89.1\pm 1.2$ & $87.5\pm 1.7$ & $85.3\pm 1$ & $78.3\pm 1.2$ \\ 
        Non-convergent & 18\,\%       & 29\,\%      & 20\,\%      & 15\,\%      & 18\,\% \\ 

    \end{tabular}
    \caption{ Results of compressed \acp{CNN} on MNIST dataset with small amount of parameters, binary weights and activations. Accuracy mean with standard deviation over normal fraction of runs is given in percent for converged runs. A fraction of training runs do not reach accuracy of 70\,\% after 10 epochs and are defined as stuck/non-convergent.}
    \label{tab:mnist_results}
\end{table}

The following investigations into detector triggering are also based on this extreme compression approach, but are evaluated using application-specific waveform data rather than image classification.

\FloatBarrier

\subsubsection{SiPM Pulse Classification\label{sec:res-sipm}}
The Search for Hidden Particles (SHiP) experiment aims to detect particles interacting feebly with ordinary matter~\cite{cernShipwebcernch}. In some of the experiment's detectors, data is generated by photo-sensitive \acf{SiPM} that convert single photons into small electrical signals.
Real-time filtering of digitized SiPM signals is essential to reduce transmitted data volume while preserving relevant detector signals. To this end, we implemented \texttt{FINN}-based 2D \acp{CNN}, \texttt{hls4ml} \acp{CNN} as conventional baselines, and two \ac{LUT}-based 2-bit \ac{BNN} variants.

All models are targeting a ZCU104 \ac{FPGA} with 
230k \acp{LUT}, 640k \acp{FF}, 1,728 \acp{DSP}, 624 (18k)\acp{BRAM} \cite{ZCU104}.

In our case, the \texttt{FINN} workflow uses the \texttt{Brevitas} framework to quantize \ac{CNN}. 
The FINN model is a 3-layer 1D-style QuantConv2d network (channels $1\text- 4\text- 6\text- 8$, kernel $5\times 1$, padding $2\times 0$, stride $1$, $8$-bit weights/activations, bias enabled) followed by an $8$-bit-activation QuantLinear classifier ($1024\to 2$, $8$-bit weights), totaling 2452 parameters.
The input signal has a frame length of $128$ samples, with $12$ bits per sample. 

The \texttt{hls4ml} workflow used the same dataset and model structure but a true 1D convolutional (Conv1d) layers trained in \texttt{PyTorch}. 
Precision is fixed to $8$-bit only at \ac{HLS}-conversion time via the \texttt{hls4ml} configuration.

The trained model was converted into \ac{HLS}, which was then synthesized into \ac{FPGA} module using the \texttt{Vitis\_HLS} tool. The generated module is subsequently integrated and verified using a testbench for functional validation and an example design for timing validation. The resulting inference latency achievable with \ac{QAT} or \ac{PTQ} techniques was higher than few $\mu$s and would be limiting the throughput of the design without further parallelization.

To overcome these limitations, we implemented \acp{BNN} trained via \acp{GA} to navigate the non-differentiable search space inherent in ultra-low-bit structures. In this architecture, weights and activations are constrained to 2-bit representations, allowing complex multiplication operations to be replaced by efficient \ac{LUT} operations within a $4 \times 4$ \ac{CAM}. 
The training process utilizes the \texttt{eaSimple} procedure with elitism, evolving a population of neural networks through mutation and crossover to maximize a fitness function derived from simulated air-shower pulses and measured noise traces. 
This evolutionary approach further optimizes the network by while maximizing accuracy also simultaneously increasing network sparsity with lower priority.

For deployment, the resulting models are synthesized through the \acs{HCL4BNN} framework, which maps the logic directly to \ac{FPGA} fabric using asynchronous combinatorial primitives such as \acp{LUT}, carry chains and multiplexers. By avoiding clocked \ac{DSP} and \ac{BRAM} operations, this hardware-constrained learning approach achieves nanosecond-scale inference latencies while significantly reducing the hardware footprint.

\begin{table}[!b]
    \centering
    \begin{tabular}{l|rr|cccc|c}
       Model  & Accuracy  & Latency & \acp{LUT} & \acsp{FF} & \acp{DSP} & \acp{BRAM}  & Training Time\\
         & \footnotesize in \% & \footnotesize in ns & \footnotesize $\times 10^3$ & \footnotesize $\times 10^3$  &  & \footnotesize 18\,k & \footnotesize min $\times$ cores \\
\hline
\texttt{FINN}     & $74\pm 4$  & 24850 & 30  & 20 & 106  & 5 & $\sim 2  \times 16$\\
\texttt{hls4ml}   &  $93\pm 2$ & 3050 & 186 & 112 & 556 & 120 & $\sim 2  \times 16$\\
BNN$_a$      &$63 \pm 3$ & 15 & 58 & 1.5 & 0 & 0 &  $\sim 300  \times 90$\\
BNN$_b$      &$74 \pm 5$ & 10 & 23 & 1.5 & 0 & 0 &  $\sim 105  \times 90$\\
BNN$_b$ (static)      &$72 \pm 3$ &  &  &  &  &  & \\
BNN$_c$      &$64 \pm 8$ & 10  & 18 & 1.5 & 0 & 0 &  $\sim 60  \times 90$
    \end{tabular}

    \caption{   
    Results and comparison of \texttt{FINN} 2DCNN ($(128 \text- 4\text- 6\text-8\text-2)$ , Kernel [5,1], padding 2.0, int8) implementation, \texttt{hls4ml} CNN, and BNN$_a$ $(128 \text- 64\text-128\text-2)$,    BNN$_b$ $(128 \text- 32\text-32\text-2)$,     BNN$_c$ $(128 \text- 16\text-64\text-2)$     with INT7 input quantization.  BNN$_b$ is also run with a static training dataset. 
    }
    \label{tab:results}
\end{table}

Three alternative BNN models are tried: BNN$_a$, BNN$_b$, and BNN$_c$ with $[16,64]$, $[32,32]$, and $[64,128]$ neurons in two hidden layers respectively, leaving the input (128) and output (2) widths same. 
With this a short-hand for architecture, \textit{e.g.} $(128\text-16\text-32\text-2)$ is used in captions.

BNN$_a$ contains 4.06 kB of tunable parameters, BNN$_b$ 1.27 kB, and BNN$_c$ 0.78 kB. The \texttt{FINN} model results in 2.4 kB parameter space, while the \texttt{hls4ml} one has 2.39 kB. 
Since \ac{BNN} weights encode combinatorial logic operations rather than multiplicative coefficients, a comparison based on matching layer topology or channel width across the three implementations would not yield a fairer comparison.
We therefore match the comparison on task, dataset, and overall parameter budget rather than architecture, and report resource and latency figures (Table~\ref{tab:results}) for judging the trade-off directly. A resource-matched (\textit{e.g.}, iso-LUT or iso-latency) comparison across paradigms is left for future work.

\begin{table}[t]
\centering
\begin{tabular}{ll|cc}
&\multicolumn{1}{c}{} &  \multicolumn{2}{c}{\text{Predicted}} \\
&  & \textbf{Good} & \textbf{Ugly}  \\
\cmidrule{2-4}
\multirow{2}{2mm}{\rotatebox[origin=c]{90}{True}} & \textbf{Good}   & 947 & 53  \\
&\textbf{Ugly}   & 410 & 590  \\
\end{tabular}
\quad\quad
\begin{tabular}{ll|cc}
&\multicolumn{1}{c}{} &  \multicolumn{2}{c}{\text{Predicted}} \\
&  & \textbf{Good} & \textbf{Ugly}  \\
\cmidrule{2-4}
\multirow{2}{2mm}{\rotatebox[origin=c]{90}{True}} & \textbf{Good}  & 984 & 16  \\
&\textbf{Ugly}  & 86 & 914  \\
\end{tabular}

\caption{ Examples of confusion matrices for \texttt{FINN} (left) and \texttt{hls4ml} (right)  for small validation sample. For the accuracies averaged over 30 runs see \ref{tab:results}.}
\label{tab:cm:baseline}
\end{table}

Since an abstention prediction is explicit for \ac{BNN} an out-of-distribution (OOD) detection test is preformed, where a random vector of input size and bit-width is evaluated by the network. 

In about 74\,\% of trainings the noise is predicted mostly as "ugly" (with less than 0.1\,\% misclassified as "good" and 0.4\,\% as "either"), about 22\,\% mostly as "either" (with less than 0.1\,\% misclassified as "good" and 0.8\,\% as "ugly"). Although, this is not evenly distributed: most networks produced no false positives at all, while a small number of runs accounted for the majority of the (still rare) misclassifications.
In 4\,\% of cases noise is classified as "good" with similar purity. Such runs should be discarded even if they have similar accuracy of predicting the main "good" and "ugly" classifiers.
It is thinkable to add noise samples to the training to obtain more robust networks, and should be studied further.
Example confusion matrices from a validation dataset are shown in Tables~\ref{tab:cm:baseline} and \ref{tab:cm:bnn}. 

\begin{table}[t]
\centering
\begin{tabular}{ll|cc|c}
&\multicolumn{1}{c}{} &  \multicolumn{3}{c}{\text{Predicted}} \\
&  & \textbf{Good} & \textbf{Ugly}  & \textbf{Either}  \\
\cmidrule{2-5}
\multirow{2}{2mm}{\rotatebox[origin=c]{90}{True}} & \textbf{Good}  & 3052 & 1948 & 0 \\
& \textbf{Ugly}  & 1595 & 3400 & 5  \\
\cmidrule{2-5}
& \textbf{Noise} &    0 &    0 & 5000 \\
\end{tabular}
\quad\quad
\begin{tabular}{ll|cc|c}
&\multicolumn{1}{c}{} &  \multicolumn{3}{c}{\text{Predicted}} \\
&  & \textbf{Good} & \textbf{Ugly}  & \textbf{Either}  \\
\cmidrule{2-5}
\multirow{2}{2mm}{\rotatebox[origin=c]{90}{True}} & \textbf{Good}  & 4883 & 117 & 0 \\
& \textbf{Ugly}  & 1533 & 3430 & 37  \\
\cmidrule{2-5}
& \textbf{Noise} &    0 &    4999 & 1 \\
\end{tabular}

\caption{ 
Example confusion matrices for one BNN$_a$  $128\text-64\text-128\text-2$ (left) and one BNN$_b$ $128\text-32\text-32\text-2$ (right);
 Either (or undecided) encodes $(1,1)$  or $(0,0)$ tuples for prediction and random input value test for truth.
}
\label{tab:cm:bnn}
\end{table}

\FloatBarrier

The \ac{LUT}-based \ac{BNN} achieves low inference latencies around $\SIrange{10}{15}{\nano\second}$ while requiring no \acp{DSP} or \acp{BRAM} (see Eq. \ref{tab:results}). 

Although classification accuracy is reduced compared to \texttt{hls4ml} implementations, the resource efficiency makes \acp{BNN} attractive for first-stage filtering under strict power budgets.

\subsubsection{Static vs. Regenerated Training Datasets for \ac{GA} \label{sec:static_regen}}
\begin{figure}[ht]
    \centering
    \includegraphics[width=0.95\linewidth]{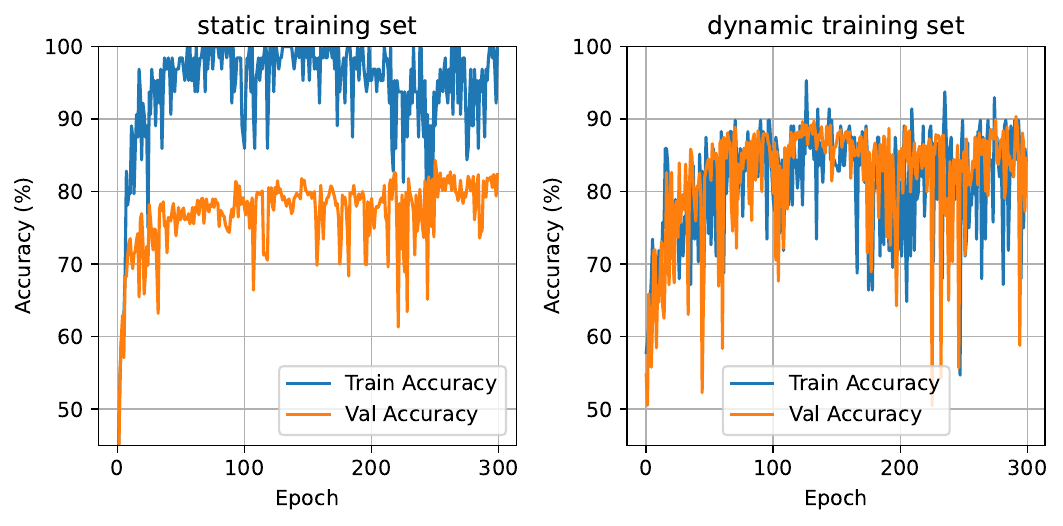}
    
    \caption{Illustration of the difference for static (left) and dynamic (right) SiPM training datasets of large (21k 4bit parameters) compressed \acp{CNN}. The training dataset here is 32+32 waveforms and 640+640 for validation. Static training data leads to exaggerated over-fitting in this illustration.}
    \label{fig:stat_dyn}
\end{figure}

To assess the effect of regenerating the training set at each fitness evaluation, we performed a dedicated control run for BNN$_b$ using a single, seed-controlled, fixed set of 200 waveforms per class ("good" and "ugly") throughout training, rather than resampling a new set each generation. 
The result is added to the Table~\ref{tab:results}, showing similar performance within uncertainty.
This is consistent with the expected effect of training on a fixed dataset: without regeneration, the \ac{GA} can partially over-fit to the specific set of training waveforms, whereas continual resampling acts akin to unlimited data augmentation and reduces this risk. Essentially the accuracy noise for the latter most strongly depends on the size of the set (smaller for training, larger for validation).
An effect is visible in Figure~\ref{fig:stat_dyn} where the network seeing validation data and dynamically regenerated training data shows similar accuracy (right), while for static training data the over-fitting is visible as discrepancy between the two (left).
For the \ac{GA} optimization, where the inference of multiple individuals dominates the computation time, time needed for the regeneration is negligible, while for backpropagation-based optimization the data regeneration has a significant impact on the overall processing time.

\FloatBarrier

\section{Discussion of Quantization Trade-offs}

Our results confirm that quantization is a key enabler for the deployment of neural networks within energy-restricted scientific edge environments, such as cryostats for quantum computing or distributed detector arrays. The investigation across different architectures and scientific domains leads to the following key conclusions:

\begin{enumerate} 
\item \textbf{\ac{PTQ}} provides a rapid and practical pathway for compressing existing models with minimal implementation effort. By reducing precision from 32-bit floating-point to 8-bit integers, we achieve a four-fold reduction in memory usage while maintaining detection quality. 

\item \textbf{\ac{QAT}} requires additional training in contrast to \ac{PTQ} and, thus, achieves slightly better results for the evaluated use case. For small neural network architectures, which are easy to train, this can be beneficial. However, for large network architectures with time-consuming training, \ac{PTQ} might be easier to apply and achieves comparable results when the quantization aims at 8-bit integers. A more aggressive quantization might yield a stronger benefit from \ac{QAT} but this has not yet been investigated.

\item \textbf{\acp{BNN}} offer an unmatched hardware efficiency, particularly for \ac{FPGA}-based inference. By constraining weights and activations to 1--2 bits, computationally intensive \ac{MAC} operations are replaced by efficient \ac{LUT} operations. This allows for nanosecond-scale latencies in the range of around $\SIrange{10}{15}{\nano\second}$ and the elimination of specialized \ac{DSP} or \ac{BRAM} blocks, which is critical for the first-level triggers in experiments like SHiP.
\end{enumerate}

Several techniques are still to be investigated in order to improve the efficiency of the \ac{BNN} approach. Most prominently, frame data, \textit{i.e.}\ short sequences, is typically stored in \ac{BRAM} structures, which have limited port width for the retrieval. On the other hand, a typical neural network input is expecting the whole dataset to perform a classification.
This creates an access bottleneck and implementing a segment-wise evaluation with several simpler networks allows for faster classifications, when a decision can be made before the the entire frame has been received. 

Moreover, \ac{GA}-based training of \acp{BNN} could incorporate a history-informed mutation bias for a fraction of operations: by tracking whether increasing or decreasing a given weight's operation code has previously correlated with fitness improvements across the population, mutation could be biased toward the empirically favorable direction rather than applied uniformly at random. This is conceptually related to estimation-of-distribution or self-adaptive mutation strategies, and could smooth the \ac{GA}'s exploration behavior. This remains speculative and untested at present, but the option only exists because of the extra state granularity of the 2-bit encoding, unlike a 1-bit weight which has no intermediate state to traverse.

Importantly, accuracy degradation is application-dependent and is often acceptable for early-stage filtering or triggering tasks where the primary objective is the significant reduction of data rates. For instance, at the Pierre Auger Observatory, AI-based triggers achieved a signal efficiency of 68\,\% compared to just 16.8\,\% for traditional methods, even after quantization to 13 bits \cite{Dorosti2025}.


\paragraph{Outlook: Cosmic-Ray Radio Detection\label{sec:res-radio}}
Another relevant use case for aggressive quantization is in the field of radio detection of cosmic-ray-induced \acp{EAS}. The target signals are short radio pulses embedded in a complex and time-varying background dominated by transient noise and \ac{RFI}. This makes simple threshold-based triggering extremely challenging: thresholds low enough to retain weak air-shower signals can lead to unacceptable false-positive rates while stricter thresholds suppress relevant events. The same background complexity also complicates the training of machine-learning triggers, since the model must distinguish rare air-shower pulses from a broad range of non-stationary backgrounds.

The Pierre Auger Observatory is one prominent experimental setting where this problem appears in practice, using large-scale radio antenna arrays to detect \acp{EAS} induced by ultra-high-energy cosmic rays. In this context, autonomous self-triggering is especially important because external triggers from particle detectors can limit the detection of highly inclined events, where the particle cascade is largely absorbed in the atmosphere while the radio signal remains measurable.


For the methodology considered here, radio traces can be represented in a waveform format comparable to the \ac{SiPM} case, using 128 samples per frame, with 12-bit signed integer amplitudes.

Previous work on ML-driven triggerring achieved a signal efficiency of approximately 68\,\% at a false-positive rate of $10^{-4}$, demonstrating the potential of learned triggers in high-interference environments \cite{Dorosti2025}.
Real-time neural network inference is therefore a promising route toward autonomous self-triggering of the radio arrays. Applying the \acp{BNN} approach studied above could further reduce the previously achieved latency of approximately \SI{2}{\micro\second}.

\section{Conclusion}
This work demonstrates that \acp{QNN} serve as a critical bridge between the high-performance requirements of modern \acp{ANN} and the severe resource constraint environments of experimental physics. The constraints in the particular environment have to be considered, as they determine the approaches to be taken.

Our results indicate that \ac{PTQ} provides a rapid and practical pathway for model compression, when latency is not of major concern. It achieves substantial reductions in memory usage ($4\times$) and computational cost ($16\times$) for 8-bit integer representations while maintaining detection quality. Using the same quantization strength, the advantage of \ac{QAT} turned out to be minimal in our use case of quantum dot calibration. For the two models investigated, a UNet-447 and a UNet-38k, the prediction quality only increases from 83.5\,\% to 84.6\,\% and from 96.0\,\% to 96.1\,\%, respectively, with a cost of additional training steps.




For applications demanding extreme energy efficiency and ultra-low latency, \acp{BNN} implemented via \acp{LUT} on \ac{FPGA} hardware represent a promising solution due to nanosecond-scale inference latencies. Our results in the use case of \ac{SiPM} readout yield to $\SIrange{10}{15}{\nano\second}$ latency without requiring specialized \ac{DSP} or \ac{BRAM} resources, which are often the limiting factor in \ac{FPGA} designs. The primary task at this stage in the readout is the reduction of irrelevant data. With a prediction quality of around 74\,\% for irrelevance ("ugly") in the most extreme case of only using $\sim 1000$ parameters, the data transmission is reduced significantly.

Also the training of \acp{BNN} has been investigated based on a \acf{GA} for our hardware-constrained, \ac{LUT}-native \ac{BNN} architecture. With this, we could extend prior work to a forward-evaluation-only 2-bit weight/activation scheme, allowing the training of non-differentiable network architectures.
The open-source \texttt{HCL4BNN} framework supports reproducible hardware-constrained learning by utilizing \acp{GA} to navigate non-differentiable search spaces. This framework successfully bridges the gap between Python-based experimentation and \ac{VHDL}-based hardware synthesis, enabling the creation of autonomous, intelligent scientific instrumentation at the edge.

Collectively, these findings contribute to a new generation of autonomous, intelligent scientific instrumentation. By integrating hardware-embedded AI directly at the edge, experimental systems can achieve real-time, on-device data processing, thereby enhancing the scalability and scientific reach of next-generation detectors.

\newpage
\section{Appendix}

\subsection{Hyperparameters Used to Train the Qubit Tuning Approach}
\label{app:unet_hyperparameters}
For training of the \acp{U-Net} that are evaluated for the use case of charge transition detection in charge stability diagrams, the PyTorch implementation of the AdamW optimizer~\cite{adamW_optimizer} is used with the parameters \texttt{lr = 0.1} and \texttt{weight\_decay = 0.0001}. Also the \texttt{OneCycleLR} scheduler is used to adjust the learning rate during training. The unquantized versions are trained for 4 epochs with a total of 1.000.000 charge stability diagrams and a batch size of 128.
For the implementation of \ac{PTQ} and \ac{QAT} the default qconfigs for the \texttt{qnnpack} backend given in PyTorch are used. For the training of the \ac{QAT} versions the quantization parameters (zero point and scale) are fixed after 3 epochs and another 3 epochs are executed to fine tune the weights with fixed quantization parameters.

\subsection{Hyperparameters of Genetic Algorithm Runs}
Table~\ref{tab:ga_hyperparams} lists the hyperparameters used for \ac{GA} training with dynamic and fixed random seeds for training and validation dataset generation.

\begin{table}[!h]
    \centering
    \begin{tabular}{l|c|p{7cm}}
        Parameter & Value & Description \\
        \hline
        \texttt{pop\_size} & 300 & Number of individuals per generation \\
        \texttt{ngen} & 50 & Number of generations the \ac{GA} is evolved for = stopping criterion \\
        \texttt{nmutbit} & 50 & Expected number of mutated bits per genome (sets per-bit mutation probability, see Sec.~\ref{sec:ga_train}) \\
        \texttt{tourn\_size} & 5 & Tournament size used for parent selection \\
        \texttt{cxpb} & 0.5 & Crossover probability per individual \\
        \texttt{cxpb\_bit} & 1 & Crossover probability per bit, given that crossover occurs \\
        \texttt{elite\_size} & 2 & Number of top individuals carried over unchanged (elitism) \\
        \hline
        \texttt{train\_n\_frames} & 300 & Number of "good"/"ugly" waveforms regenerated per fitness evaluation during training \\
        \texttt{train\_static\_seed} & 123 & Random seed fixed for the training data generator \\
        \texttt{validation\_n\_frames} & 5000 & Number of waveforms used for the independent validation/test evaluation \\
        \texttt{validation\_static\_seed} & 1 & Random seed fixed for the validation data generator \\
    \end{tabular}
    \caption{Genetic algorithm hyperparameters and dataset generation settings used for \ac{BNN} training.}
    \label{tab:ga_hyperparams}
\end{table}

\newpage
\subsection{Glossary of Relevant Hardware-Specific Expressions}
\label{sec:glossary}
\paragraph{FPGA} A \acf{FPGA} is a programmable integrated circuit for logic operations. It consists of repeating elements, especially \acp{LUT}, \acp{DSP}, and memory blocks. By combining these fundamental elements together, logic functions can be programmed onto the \ac{FPGA}. The instructions are usually given in a \ac{HDL} and then synthesized for the specific hardware.

\paragraph{LUT} A \acf{LUT} is a fundamental building block of an \ac{FPGA}. It provides a configurable function based on full adders and single bit storage. 

\paragraph{DSP} A \acf{DSP} is a larger fundamental building block inside of an \ac{FPGA} that allows more complex signal operations, for instance multiplications. It is a much scarcer resource than \acp{LUT} on \acp{FPGA}.

\paragraph{BRAM} Data storage in an \ac{FPGA} depends on the amount of data and accessible hardware components. Usually available inside the \ac{FPGA} fabric and most flexible is storing data in a memory segment called \acf{BRAM}. Denser and more efficient storage may be available with external components but is not part of the fabric and requires dedicated memory controllers.

\paragraph{HDL} A \acf{HDL} is a hardware-compatible description of the logic that can be ported onto integrated circuits. Similar to programming languages, the instructions have to be converted to fundamental instructions that are compatible with hardware. The major difference to programming languages is that \ac{HDL} describes the connection of physical systems and electrical signals. Common languages are \textbf{\acs{VHDL}} and \textbf{Verilog}.

\paragraph{HLS} The hardware description in a \ac{HDL} can be done manually or through an automized process. By using a \acf{HLS}, the user describes the intended function in typical programming language like C++ and converts it via a framework into \ac{HDL}.

\subsection{Declarations}

\begin{data}
Source code was published in Zenodo under \cite{HCL4BNN}. SiPM training data for \acp{BNN} was generated by this code using the \texttt{SiPMDataset} Python class.
\end{data}

\begin{aiuse}
\ac{LLM} tools were used solely for linguistic editing, including wording, grammar, and style. 
All scientific content, analyses, interpretations, and conclusions were developed by the authors. 
All code for the framework was authored and reviewed by humans, and all changes were merged only after human review and verification.
\end{aiuse}






\null
\newpage
\bibliographystyle{unsrt}
\bibliography{article}

\end{document}